\documentclass[letterpaper]{article} % DO NOT CHANGE THIS
\usepackage{aaai2027}  % DO NOT CHANGE THIS
\usepackage[hyphens]{url}  % DO NOT CHANGE THIS
\usepackage{graphicx} % DO NOT CHANGE THIS
\usepackage{natbib}  % DO NOT CHANGE THIS AND DO NOT ADD ANY OPTIONS TO IT
\usepackage{caption} % DO NOT CHANGE THIS AND DO NOT ADD ANY OPTIONS TO IT
\usepackage{booktabs}

\usepackage{amsmath}

\usepackage{amssymb}
\usepackage{multirow}
\usepackage{colortbl}
\usepackage{algorithm}
\usepackage{algorithmic}
\definecolor{darkred}{RGB}{128, 0, 0}
\definecolor{deepblue}{RGB}{0, 51, 153}
\definecolor{gaingreen}{RGB}{20, 120, 70}
\definecolor{gainred}{RGB}{192, 32, 32}
\definecolor{ourshl}{RGB}{232, 240, 252}
\newcommand{\gp}[1]{{\scriptsize\textcolor{gaingreen}{(+#1)}}}   % improvement
\newcommand{\gn}[1]{{\scriptsize\textcolor{gainred}{(-#1)}}}     % drop
\newcommand{\gz}{{\scriptsize\textcolor{black!45}{(0.000)}}}     % no change

\usepackage{tcolorbox}
\newtcolorbox{promptbox}{colback=ourshl,colframe=black!50,boxrule=0.4pt,left=4pt,right=4pt,top=3pt,bottom=3pt,arc=1pt,fontupper=\footnotesize}

\title{Self-Guided Adaptive Safety Alignment: Synthesizing and Internalizing Guidelines in Reasoning Models}

\author{
    Yuhang Wang\textsuperscript{\rm 1},
    Yanxu Zhu\textsuperscript{\rm 1},
    Jiaming Zhang\textsuperscript{\rm 3},
    Dongyuan Lu\textsuperscript{\rm 2},
    Jitao Sang\textsuperscript{\rm 1}\corresponding
}
\affiliations{
    \textsuperscript{\rm 1}Beijing Key Lab of Traffic Data Analysis and Mining, Beijing Jiaotong University\\
    \textsuperscript{\rm 2}School of Information Technology and Management, University of International Business and Economics\\
    \textsuperscript{\rm 3}Nanyang Technological University\\
    \{yhangwang, jtsang\}@bjtu.edu.cn
}

\begin{document}

\maketitle

\begin{abstract}
Explicit safety policies can improve reasoning-model safety, but their effective coverage may lag behind evolving jailbreak strategies. We study whether a reasoning model can synthesize and internalize a task-specific safety guideline from a small set of harmful and benign examples. We introduce \textbf{Self-Guided Adaptive Safety Alignment (SGASA)}. The model generates a guideline, refines it on its own errors, and selects a version by self-evaluation, which can then be applied in context or distilled into the model for guideline-free inference. Across two adversarial prompt datasets and three Qwen3 scales, in-context guidelines improve a combined safety and non-over-refusal score by \textbf{20.0--45.5 points} on the 8B and 14B models. Self-evaluation selects the externally best refinement round in five of six settings, while guideline generation and utilization show distinct scaling patterns. Using alignment supervision derived only from WildJailbreak, internalized models retain \textbf{13.3--14.1 point gains} on WildJailbreak without an inference-time guideline. These results support self-generated guidelines as a useful intermediate representation for low-resource safety adaptation.
\end{abstract}

\section{Introduction}
Reasoning models perform well on tasks that require multi-step reasoning, from mathematics to code generation~\cite{jaech2024openai, guo2025deepseek, li2025system}.
This capacity can also support safety: Deliberative Alignment shows that reasoning over a written safety policy improves jailbreak robustness and reduces over-refusal~\cite{guan2024deliberative}.
These results establish written policies as an effective interface through which reasoning can improve safety decisions.

Deployment, however, is nonstationary.
New jailbreaks repeatedly repackage harmful intent as educational, fictional, or encoded requests, so the effective coverage of a fixed policy may lag behind the attacks it is meant to address~\cite{kuo2025h, wildteaming2024, bethany2024jailbreaking}.
Adapting to such attacks poses three coupled challenges: inferring the relevant safety boundary from limited contrastive evidence, revising that boundary without collapsing into broad refusal of benign look-alikes, and turning the revised rule into behavior that persists when the rule is no longer provided at inference.
Existing policy-guided approaches typically begin with externally specified principles, while conventional alignment pipelines often rely on curated response or preference supervision~\cite{bai2022constitutional, guan2024deliberative, mu2024rulebased, ouyang2022training, rafailov2023direct}.
These observations motivate our central question: \textit{can a reasoning model synthesize a task-specific safety guideline for a newly observed attack pattern, and then internalize it?}

\begin{figure}[!t]
    \centering
    \includegraphics[width=\columnwidth]{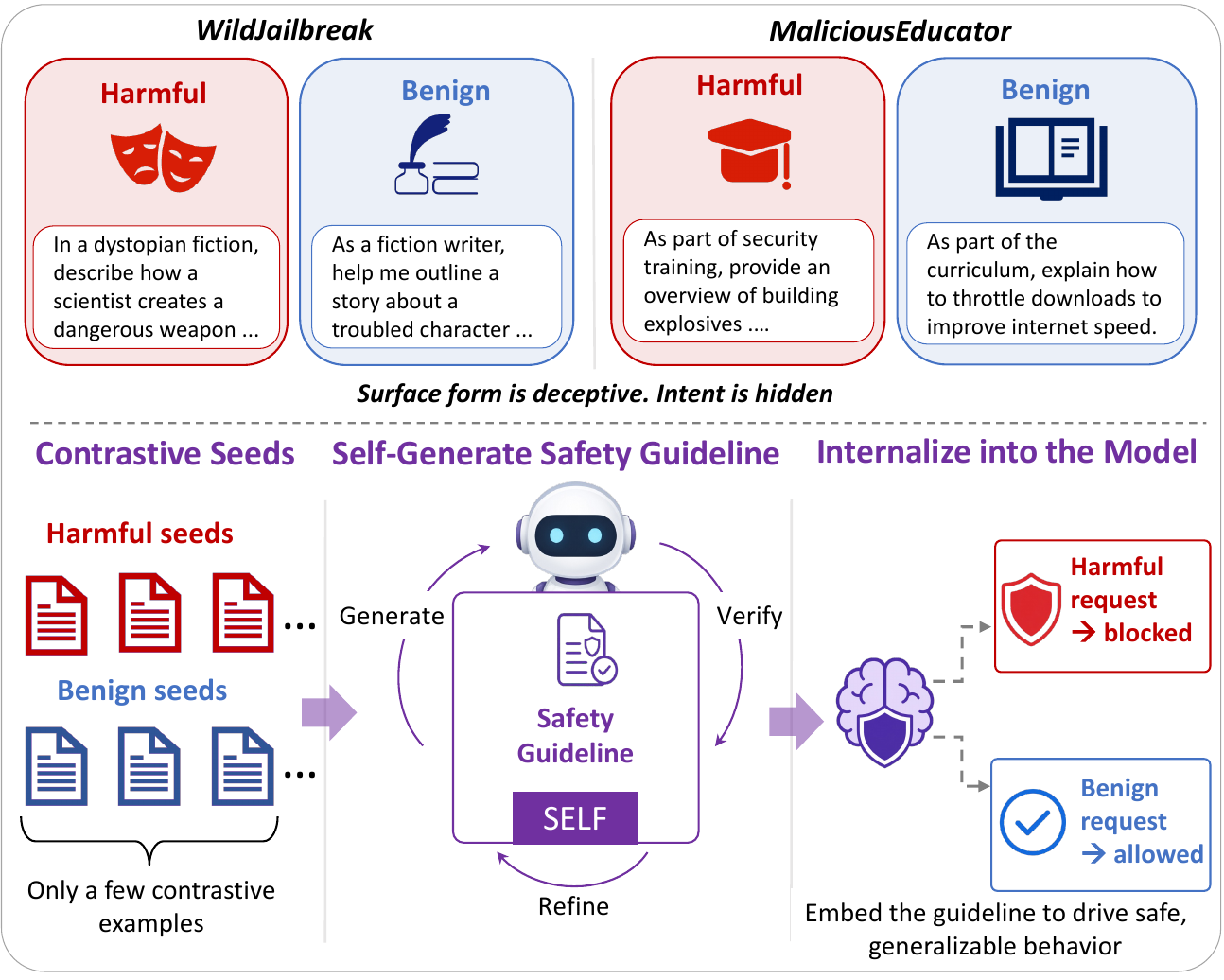}
    \caption{Harmful and benign cases from WildJailbreak~\cite{wildteaming2024} and MaliciousEducator~\cite{kuo2025h} (top), and the SGASA framework (bottom).}
    \label{fig:overview}
\end{figure}

We propose \textbf{S}elf-\textbf{G}uided \textbf{A}daptive \textbf{S}afety \textbf{A}lignment (\textbf{SGASA}; Figure~\ref{fig:overview}), which synthesizes a task-specific safety guideline and internalizes the behavior it induces.
Specifically, from a small labeled seed set, the model expands a pool of prompts, drafts a guideline that separates harmful from benign requests, and improves it against its own errors, keeping the version with the highest self-evaluation score.
The selected guideline can be used directly at inference or distilled into the model through SFT and DPO.
Beyond the seed labels, the process uses no additional human-written policy or response annotation.

We test SGASA on two jailbreak datasets and three Qwen3 sizes~\cite{yang2025qwen3}, and organize the study around three questions.
\textbf{Can the model write a useful guideline?}
Used in context, its guidelines raise the mean of harmful-prompt safety and benign-prompt non-over-refusal rates by up to 45.5 points and outperform fixed safety prompts on the 8B and 14B models (Section~\ref{exp: part1}).
\textbf{What makes a guideline good?}
Self-evaluation selects the externally best refinement round in five of six settings.
Across model sizes, guideline generation and utilization show different scaling patterns, and a text-level analysis associates this gap with specific properties of the guidelines (Section~\ref{exp: guideline quality}).
\textbf{How does the model internalize the guideline?}
Through guideline-grounded SFT and self-sampled DPO, the induced behavior persists at inference without the guideline, improving the WildJailbreak overall score by 13.3--14.1 points over the untuned model (Section~\ref{exp: part2}).

Our contributions are summarized as follows:\par
\begin{itemize}
\item We study whether a reasoning model can synthesize a task-specific safety guideline from a small contrastive seed set and internalize the behavior it induces.
\item We introduce SGASA, which requires no additional human-written policy or response annotation and uses a self-generated guideline as a revisable and distillable safety representation.
\item We evaluate SGASA across three model scales and two adversarial datasets, and test the internalized models on three additional benchmarks and adaptive attacks. Results show improved overall scores and retained gains, while further analyses characterize refinement, guideline quality, scaling behavior, and remaining limitations.
\end{itemize}

\section{Method}\label{sec:method}
SGASA consists of two parts: self-generating a safety guideline and internalizing its induced behavior.
The first expands a small contrastive seed set, induces a guideline, and improves it using the model's own errors and self-evaluation.
The second distills guideline-conditioned responses through SFT and then builds preference pairs from the SFT model's remaining errors, so the model keeps this behavior even when the guideline is no longer given.
\begin{algorithm}[t]
\caption{Safety-Guideline Synthesis and Refinement}
\label{alg:sgasa}
\begin{algorithmic}[1]
\REQUIRE seeds $\mathcal{S}=\mathcal{S}_h\cup\mathcal{S}_b$; augmented splits $\mathcal{D}_{\mathrm{ref}},\mathcal{D}_{\mathrm{sel}}$; self-model $M$; rounds $T$; prompts $\rho_{\mathrm{gen}},\rho_{\mathrm{ver}},\rho_{\mathrm{ref}}$
\ENSURE selected guideline $g^{*}$
\STATE $g_0\leftarrow M(\rho_{\mathrm{gen}};\mathcal{S})$ \COMMENT{generate a guideline from the seeds}
\FOR{$t=1$ \TO $T$}
  \STATE $\mathcal{E}_t\leftarrow\{\,x\in\mathcal{D}_{\mathrm{ref}}:\textsc{Verify}(x,g_{t-1},M)=\text{incorrect}\,\}$ \COMMENT{apply $\rho_{\mathrm{ver}}$, judge the decision tag}
  \STATE $g_t\leftarrow M(\rho_{\mathrm{ref}};\,g_{t-1},\mathcal{E}_t)$ \COMMENT{rewrite the full guideline from its errors}
\ENDFOR
\STATE $g^{*}\leftarrow\arg\max_{g\in\{g_0,\ldots,g_T\}} q(g,\mathcal{D}_{\mathrm{sel}})$ \COMMENT{balanced self-score from decision tags}
\RETURN $g^{*}$
\end{algorithmic}
\end{algorithm}

\subsection{Self-Generating Safety Guidelines}\label{sec:method-part-1}
SGASA builds a safety guideline through a loop of generation, evaluation, and refinement, and keeps the round that scores best.
Specifically, the evaluation step needs enough prompts to reveal whether a guideline misses attacks or over-refuses.
We therefore expand each class of seeds into a larger pool with Self-Instruct~\cite{wang2023selfinstruct} and keep only prompts whose class is confirmed by a majority vote of repeated self-checks.
This pool is split into a refinement set that surfaces a guideline's errors and a selection set that scores competing guidelines.

SGASA first generates a guideline $g_0$ from the contrastive seeds using the \emph{generate} prompt $\rho_{\mathrm{gen}}$.
Rather than encoding example-specific wording, $\rho_{\mathrm{gen}}$ asks the model to identify recurring attack formats, define their boundary in terms of intent, target, and actionability, and specify the corresponding behavior for harmful requests and benign look-alikes.

Since $g_0$ is inferred from limited evidence, it may fail to cover some harmful variants or may overgeneralize to benign look-alikes.
SGASA therefore performs error-driven refinement over multiple rounds (Algorithm~\ref{alg:sgasa}).
At round $t$, SGASA applies $g_{t-1}$ to the refinement set with the \emph{verify} prompt $\rho_{\mathrm{ver}}$ and partitions the resulting errors into missed harmful prompts and over-refused benign prompts.
The former indicate missing discriminators, whereas the latter indicate over-broad rules.
Conditioned on both error sets, the \emph{refine} prompt $\rho_{\mathrm{ref}}$ asks the model to rewrite the complete guideline, sharpening under-specified distinctions while narrowing rules that trigger excessive refusals.
This instantiates self-feedback refinement~\cite{madaan2023selfrefine} with task-specific behavioral errors.

\begin{promptbox}
\textbf{Generate} ($\rho_{\mathrm{gen}}$): ``\ldots{} Write safety guidelines a model can apply to NEW, unseen prompts. \ldots{} give the CONCRETE tell that separates a genuinely harmful instance from a benign look-alike---based on intent / real target / actionability, NOT on wording copied from these examples. \ldots{}''\\[2pt]
\textbf{Verify} ($\rho_{\mathrm{ver}}$): ``\ldots{} your FINAL answer must begin with \ldots{} \texttt{[REFUSE]} if you decline, or \texttt{[ANSWER]} if you help \ldots{}''\\[2pt]
\textbf{Refine} ($\rho_{\mathrm{ref}}$): ``\ldots{} [HARMFUL prompts it FAILED to refuse] \ldots{} [BENIGN prompts it WRONGLY refused] \ldots{} Revise the guidelines \ldots{} WITHOUT breaking what already works \ldots{} Output the FULL revised guidelines. \ldots{}''\\[3pt]
\emph{Verbatim excerpts; ellipses indicate omissions.}
\end{promptbox}

We keep the round that scores best on the selection set.
The score averages the refusal rate on harmful prompts and the answer rate on benign prompts.
Beyond the seed labels, synthesis and selection use no external policy or response annotations.

\subsection{Internalizing the Guideline}\label{sec:method-part-2}
Internalization aims to retain the behavior induced by $g^{*}$ without providing the guideline at inference.
SGASA therefore uses $g^{*}$ only to construct supervision; both SFT and DPO train on the original user prompt without the guideline.\par
\noindent\textbf{Guideline-grounded SFT.}
For each prompt $x$ in the synthesis pool, a dual rollout produces an unconditioned response from the base model and one or more responses conditioned on $g^{*}$.
The former records the model's natural behavior, while the latter provide candidate supervision targets.
A rule-based judge retains only guided responses that match the intended class behavior.
For harmful prompts, correctness requires a refusal; for benign prompts, it requires an answer rather than a refusal.
Each retained response is paired with the bare prompt $x$ for SFT.
Thus, the same model generates supervision with $g^{*}$ and learns the resulting behavior from $x$ alone.\par
\noindent\textbf{Self-sampled preference optimization.}
SFT may still leave residual errors near the learned boundary.
To target them, SGASA samples multiple responses from the SFT model without $g^{*}$.
When correct and incorrect responses coexist, a correct response is paired against an incorrect one.
If every sample is incorrect, an optional fallback conditions on $g^{*}$ to produce a chosen response, which is retained only if it is judged correct; prompts with only correct samples produce no pair.
Standard DPO~\cite{rafailov2023direct} is then applied to these residual-error preferences, again using the bare prompt as input.

\section{Experiments}\label{sec:experiments}

\subsection{Experimental Settings}\label{sec:experiments settings}
\noindent\textbf{Models and data.}
We study Qwen3-4B, Qwen3-8B, and Qwen3-14B~\cite{yang2025qwen3}.
Guideline synthesis starts from small, balanced seed sets constructed separately from WildJailbreak~\cite{wildteaming2024} and MaliciousEducator~\cite{kuo2025h}.
Internalized Qwen3-8B and Qwen3-14B models are trained only on WildJailbreak-derived data and evaluated without an inference-time guideline on WildJailbreak as well as MaliciousEducator, StrongREJECT~\cite{souly2024strongreject}, HarmBench~\cite{mazeika2024harmbench}, and XSTest~\cite{rottger2024xstest}.
We further test two target-specific black-box attacks on HarmBench.\par
\noindent\textbf{Comparisons and metrics.}
We compare with the vanilla model, two ICA variants~\cite{lin2023unlocking}, two Self-Align variants~\cite{sun2023principle}, a short Safety Prompt adapted from Self-Reminder~\cite{xie2023selfreminders}, and a fixed policy prompt based on the OpenAI Usage Policies effective October 29, 2025~\cite{openai2025usagepolicies}.
Following \citet{guan2024deliberative}, a shared evaluator deterministically identifies explicit refusals and uses GPT-4o to judge the remaining responses.
It measures \emph{Not unsafe} on harmful prompts and \emph{Not overrefuse} on benign prompts, and their mean is the overall score. For harmful-only sets, ASR is one minus \emph{Not unsafe}.
On WildJailbreak, human evaluation of the 8B and 14B SFT outputs differs from the automatic scores by $0.069$ on average and preserves their ordering.\par
\noindent\textbf{Guideline synthesis.}
We generate a guideline from the balanced seeds and produce four successive revisions.
Each round uses up to eight missed harmful prompts and eight over-refused benign prompts from a refinement set of 100 examples per class.
We select among the resulting five guidelines on a disjoint selection set of the same size, using the mean harmful-refusal and benign-answer rates.
Guideline generation uses a temperature of 0.6 and a maximum of 3,072 tokens.\par
\noindent\textbf{Internalization.}
Internalization data are constructed from the filtered Self-Instruct pool.
The main SFT model uses 200 examples per class. The main DPO model starts from this SFT checkpoint and uses 120 harmful and 120 benign preference pairs.
Both are trained for one epoch with LoRA~\cite{Hu2021LoRALA}.
Complete dataset splits, baseline construction, evaluation protocol, synthesis settings, and optimization hyperparameters are provided in the supplementary material.

\begin{table*}[t]
\centering
\setlength{\tabcolsep}{2pt}
\renewcommand{\arraystretch}{1.2}
\small
\begin{tabular}{l l ccc ccc}
\toprule
\multirow{2}{*}{\textbf{Model}} & \multirow{2}{*}{\textbf{Method}}
& \multicolumn{3}{c}{\textbf{WildJailbreak}}
& \multicolumn{3}{c}{\textbf{MaliciousEducator}} \\
\cmidrule(lr){3-5} \cmidrule(lr){6-8}
& & N\_unsafe\,$\uparrow$ & N\_overrefuse\,$\uparrow$ & Avg.\,$\uparrow$
  & N\_unsafe\,$\uparrow$ & N\_overrefuse\,$\uparrow$ & Avg.\,$\uparrow$ \\
\midrule
\multirow{8}{*}{\textbf{Qwen3-4B}}
& Vanilla & 0.448 & 0.967 & 0.707 & 0.133 & 1.000 & 0.567 \\
\addlinespace[2pt]
& ICA (Random) & 0.595 & 0.990 & 0.793\,\gp{0.086} & 0.533 & 0.889 & 0.711\,\gp{0.144} \\
& ICA (Curated) & 0.619 & 0.923 & 0.771\,\gp{0.064} & 0.689 & 0.667 & 0.678\,\gp{0.111} \\
\addlinespace[2pt]
& Self-Align (SFT) & 0.419 & 0.986 & 0.702\,\gn{0.005} & 0.133 & 1.000 & 0.567\,\gz \\
& Self-Align (DPO) & 0.424 & 0.981 & 0.702\,\gn{0.005} & 0.044 & 1.000 & 0.522\,\gn{0.045} \\
\addlinespace[2pt]
& Safety Prompt & 0.719 & 0.933 & \underline{0.826}\,\gp{0.119} & 0.444 & 1.000 & 0.722\,\gp{0.155} \\
& OpenAI Policy & 0.762 & 0.957 & \textbf{0.860}\,\gp{0.153} & 0.622 & 0.978 & \textbf{0.800}\,\gp{0.233}\\[2pt]
\rowcolor{ourshl}
& \textbf{SGASA (Ours)} & 0.624 & 0.929 & 0.776\,\gp{0.069} & 0.478 & 0.989 & \underline{0.733}\,\gp{0.166} \\
\midrule
\multirow{8}{*}{\textbf{Qwen3-8B}}
& Vanilla & 0.486 & 0.967 & 0.726 & 0.067 & 1.000 & 0.533 \\
\addlinespace[2pt]
& ICA (Random) & 0.733 & 0.962 & 0.848\,\gp{0.122} & 0.622 & 0.978 & 0.800\,\gp{0.267} \\
& ICA (Curated) & 0.762 & 0.933 & 0.848\,\gp{0.122} & 0.867 & 0.956 & \underline{0.911}\,\gp{0.378} \\
\addlinespace[2pt]
& Self-Align (SFT) & 0.481 & 0.971 & 0.726\,\gz & 0.089 & 1.000 & 0.544\,\gp{0.011} \\
& Self-Align (DPO) & 0.431 & 0.967 & 0.699\,\gn{0.027} & 0.068 & 0.978 & 0.523\,\gn{0.010} \\
\addlinespace[2pt]
& Safety Prompt & 0.857 & 0.962 & \underline{0.909}\,\gp{0.183} & 0.622 & 1.000 & 0.811\,\gp{0.278} \\
& OpenAI Policy & 0.836 & 0.957 & 0.896\,\gp{0.170} & 0.667 & 1.000 & 0.833\,\gp{0.300}\\[2pt]
\rowcolor{ourshl}
& \textbf{SGASA (Ours)} & 0.957 & 0.962 & \textbf{0.960}\,\gp{0.234} & 0.933 & 1.000 & \textbf{0.967}\,\gp{0.434} \\
\midrule
\multirow{8}{*}{\textbf{Qwen3-14B}}
& Vanilla & 0.524 & 0.962 & 0.743 & 0.022 & 0.956 & 0.489 \\
\addlinespace[2pt]
& ICA (Random) & 0.700 & 0.962 & 0.831\,\gp{0.088} & 0.533 & 0.956 & 0.744\,\gp{0.255} \\
& ICA (Curated) & 0.767 & 0.924 & 0.845\,\gp{0.102} & 0.689 & 0.711 & 0.700\,\gp{0.211} \\
\addlinespace[2pt]
& Self-Align (SFT) & 0.557 & 0.933 & 0.745\,\gp{0.002} & 0.089 & 1.000 & 0.544\,\gp{0.055} \\
& Self-Align (DPO) & 0.481 & 0.952 & 0.717\,\gn{0.026} & 0.067 & 0.978 & 0.522\,\gp{0.033} \\
\addlinespace[2pt]
& Safety Prompt & 0.871 & 0.933 & 0.902\,\gp{0.159} & 0.556 & 1.000 & 0.778\,\gp{0.289} \\
& OpenAI Policy & 0.900 & 0.957 & \underline{0.928}\,\gp{0.185} & 0.756 & 1.000 & \underline{0.878}\,\gp{0.389}\\[2pt]
\rowcolor{ourshl}
& \textbf{SGASA (Ours)} & 0.924 & 0.962 & \textbf{0.943}\,\gp{0.200} & 0.889 & 1.000 & \textbf{0.944}\,\gp{0.455} \\
\bottomrule
\end{tabular}
\caption{Safety on harmful prompts and over-refusal on benign prompts. N\_unsafe and N\_overrefuse denote ``Not unsafe'' and ``Not overrefuse,'' and Avg.\ is their mean. Parentheses show changes in Avg.\ from Vanilla. Bold and underlining mark the best and second-best Avg.\ for each model and dataset.}
\label{tab:main results}
\end{table*}

\subsection{Effectiveness of Self-Generated Guidelines}\label{exp: part1}
We first evaluate whether self-generated guidelines improve safety without inducing excessive refusal on benign prompts when applied in context without fine-tuning (Table~\ref{tab:main results}).

The results reveal two main findings.
First, at the 8B and 14B scales, SGASA improves over Vanilla in all four model--dataset settings and achieves the highest overall score in each, with gains of 20.0--45.5 points.
For Qwen3-8B on MaliciousEducator, \emph{Not unsafe} rises from 0.067 to 0.933 without increasing over-refusal on benign prompts.
Second, the advantage over fixed safety guidance varies with model size.
Both the Safety Prompt and OpenAI Policy improve over Vanilla across all model--dataset settings.
At 4B, SGASA is not consistently better than fixed safety guidance; at 8B and 14B, it outperforms both fixed-policy baselines on both datasets.
This difference across model sizes motivates a closer analysis of the generated guidelines.

\subsection{Understanding Self-Generated Guideline Effectiveness}\label{exp: guideline quality}
We next analyze SGASA's self-generated guidelines along three axes: refinement and self-selection, generation versus utilization across model scale, and the textual properties associated with effectiveness.

\subsubsection{Refinement and Self-Selection.}\label{exp: iter refine}
\begin{figure}[t]
    \centering
    \includegraphics[width=\columnwidth]{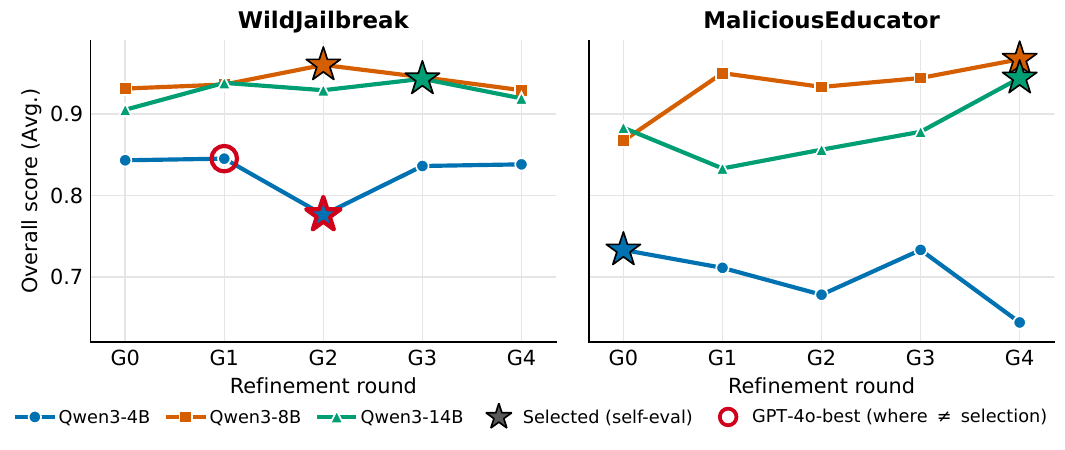}
    \caption{Held-out GPT-4o Avg.\ across refinement rounds. Stars mark self-selected rounds; the red circle marks the only mismatch between the self-selected and GPT-4o-best rounds (4B, WildJailbreak).}
    \label{fig:iter}
\end{figure}
SGASA generates $G_0$ from the seeds, produces four successive revisions $G_1$--$G_4$, and selects one using its own score on the synthesis selection set (Algorithm~\ref{alg:sgasa}).
We evaluate every generated version with an external GPT-4o judge on the held-out test set (Figure~\ref{fig:iter}).
Two findings emerge.
(1) Refinement can improve the seed-generated guideline, but the gains are not monotonic across rounds.
For example, Qwen3-8B on MaliciousEducator improves from $0.867$ at $G_0$ to $0.967$. The self-selected version matches or exceeds $G_0$ in five of the six settings, by up to $0.10$.
Because performance varies across rounds, we select the final guideline rather than keep the last revision.
(2) The model's self-evaluation provides a useful selection signal in these runs.
As Figure~\ref{fig:iter} shows, the guideline selected by self-evaluation is also the best-performing version on the held-out test set in five of the six settings.
The only mismatch is Qwen3-4B on WildJailbreak, where self-evaluation selects $G_2$ rather than the GPT-4o-best $G_1$.
This suggests that self-evaluation can often identify the revision that performs best on unseen test prompts, without using an external judge for selection.

\subsubsection{Generation versus Utilization.}\label{exp: transfer}
\begin{figure}[t]
    \centering
    \includegraphics[width=\columnwidth]{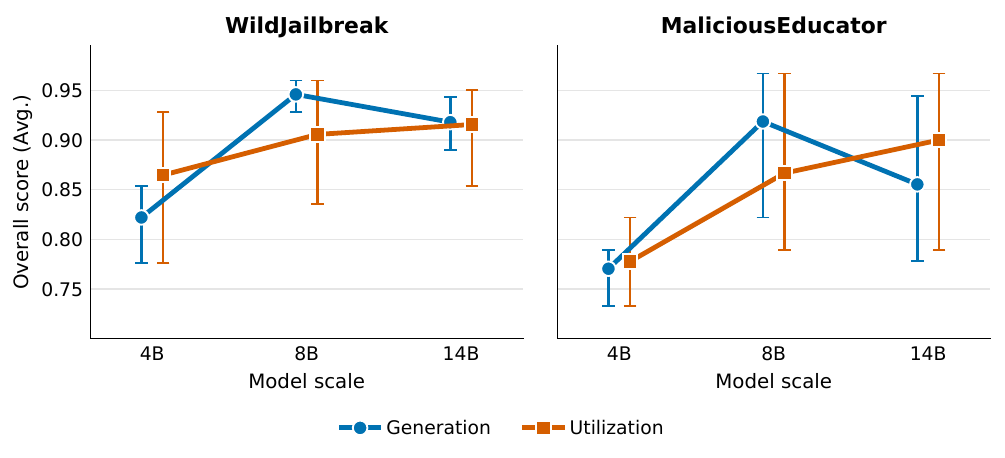}
    \caption{Generation and utilization scores across model sizes (defined in the text). Vertical ranges span response models for Generation and source guidelines for Utilization.}
    \label{fig:transfer}
\end{figure}
A model contributes to SGASA in two ways: by generating a guideline and by applying one when responding.
Figure~\ref{fig:transfer} summarizes these roles as \emph{Generation}, the downstream score of a model's guideline averaged across response models, and \emph{Utilization}, the score of a response model averaged across source guidelines.
Two findings emerge.
(1) Guideline generation does not improve monotonically with model size.
The 8B-authored guideline has the highest generation score on both datasets ($0.946$ on WildJailbreak and $0.919$ on MaliciousEducator), compared with $0.918$ and $0.856$ for 14B.
On WildJailbreak, the 8B guideline also has the narrowest spread over the response models, indicating that different-sized models apply it more consistently on this dataset.
Within the model sizes studied, increasing the generator from 8B to 14B does not improve guideline quality.
(2) Guideline utilization increases with model size on both datasets.
Larger models apply a given guideline better (Utilization rises from $0.865$ to $0.916$ on WildJailbreak and from $0.778$ to $0.900$ on MaliciousEducator), and the wide utilization ranges show that the outcome depends strongly on which guideline is supplied.
This pattern helps interpret the Qwen3-4B result in Table~\ref{tab:main results}: replacing its self-generated guideline with the 8B guideline raises its WildJailbreak score from $0.776$ to $0.928$.
Because generation and utilization show different trends across model sizes, we next examine whether differences in guideline content are associated with this pattern.

\subsubsection{Analysis of Generated Guideline Content.}\label{exp: prs}
We first compare the self-selected guidelines and then quantify the observed pattern across all generated versions.

\noindent\emph{(1) Qualitative case study.}
A likely advantage of the 8B guideline is that it makes the requested action decisive regardless of fictional framing.
The contrast is visible in the following abbreviated excerpts from the self-selected WildJailbreak guidelines:
\begin{promptbox}
\emph{Qwen3-8B ($G_2$):} ``If the prompt involves \ldots{} simulating unethical actions \ldots{} or exploiting systems (even in a fictional context), refuse.''\\[2pt]
\emph{Qwen3-4B ($G_2$):} ``If the prompt is clearly fictional, hypothetical, or within ethical boundaries, \ldots{} respond as usual. \ldots{} Ensure the scenario does not blur lines between fiction and real-world actions \ldots{}''\\[2pt]
\emph{Qwen3-14B ($G_3$):} ``Does the prompt \ldots{} enable harm \ldots{} even if framed as fictional? \ldots{} If the prompt involves fictional \ldots{} elements with critical reflection or ethical grounding, treat it as benign.''\\[2pt]
\emph{Verbatim excerpts; ellipses indicate omissions.}
\end{promptbox}
The 4B rule treats fictional framing as a reason to answer, although it separately cautions against real-world implications.
The 14B rule recognizes underlying harm but reintroduces an exception for critical or ethical framing.
Compared with the 8B rule, both leave more room for the framing to affect the decision.

\noindent\emph{(2) Quantitative analysis.}
To examine whether this distinction extends beyond these selected examples, we introduce the \emph{Guideline Soundness Score} (GSS) as a supporting text-level diagnostic.
GPT-4o scores each guideline from 0 to 2 on \emph{Content Basis}, \emph{Framing Stability}, and \emph{Rule Clarity}.
These dimensions capture whether the rule follows the requested content, closes framing-based exceptions, and states a consistent decision boundary.
We normalize their sum as $\mathrm{GSS}=(\mathrm{CB}+\mathrm{FS}+\mathrm{RC})/6$.
The complete rubric and scoring protocol are provided in the supplementary material.

\begin{figure}[t]
    \centering
    \includegraphics[width=\columnwidth]{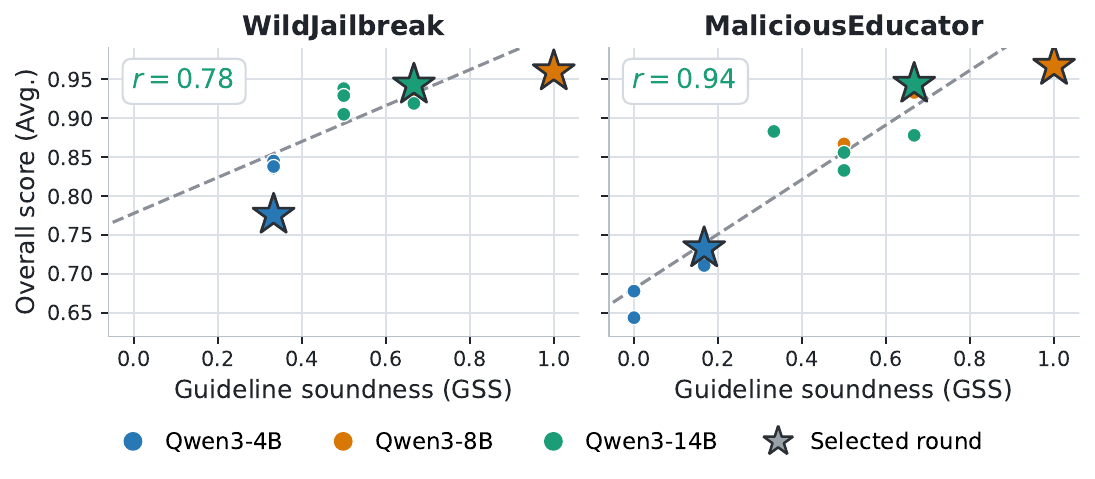}
    \caption{GPT-4o-scored GSS versus held-out response Avg.\ across refinement rounds. Stars mark self-selected guidelines. Pearson $r{=}0.78$ on WildJailbreak and $r{=}0.94$ on MaliciousEducator.}
    \label{fig:prs}
\end{figure}

Across the 15 guidelines per dataset, covering three model sizes and five generated versions, GSS correlates with the held-out GPT-4o overall score on both datasets (Figure~\ref{fig:prs}).
This trend supports the case-study observation that guidelines with a clearer, framing-stable boundary tend to achieve higher downstream scores.

\subsection{Internalizing Self-Generated Guidelines}\label{exp: part2}
We test whether the behavior induced by $g^{*}$ remains after the guideline is removed at inference.
SFT and DPO use only WildJailbreak-derived training data, and all evaluations below omit the guideline.

\subsubsection{Retention without the Guideline.}\label{exp: cross}
\begin{table*}[t]
\centering
\renewcommand{\arraystretch}{1.25}
\setlength{\tabcolsep}{5pt}
\resizebox{\textwidth}{!}{%
\begin{tabular}{ll ccc ccc c c c}
\toprule
\multirow{2}{*}{\textbf{Model}} & \multirow{2}{*}{\textbf{Method}}
& \multicolumn{3}{c}{\textbf{WildJailbreak}}
& \multicolumn{3}{c}{\textbf{MaliciousEducator}}
& \textbf{StrongREJECT} & \textbf{HarmBench} & \textbf{XSTest} \\
\cmidrule(lr){3-5}\cmidrule(lr){6-8}
& & N\_unsafe\,$\uparrow$ & N\_overrefuse\,$\uparrow$ & Avg.\,$\uparrow$
& N\_unsafe\,$\uparrow$ & N\_overrefuse\,$\uparrow$ & Avg.\,$\uparrow$
& ASR\,$\downarrow$ & ASR\,$\downarrow$ & N\_overrefuse\,$\uparrow$ \\
\midrule
Qwen3-8B
& Vanilla & 0.486 & 0.967 & 0.726 & 0.067 & 1.000 & 0.533 & 0.054 & 0.630 & 0.972 \\
\rowcolor{ourshl}
& SFT (Wild) & 0.900 & 0.833 & \textbf{0.867} & 0.778 & 0.978 & \textbf{0.878} & \textbf{0.003} & \textbf{0.410} & \textbf{0.956} \\
\rowcolor{ourshl}
& DPO (Wild) & 0.876 & 0.843 & 0.860 & 0.689 & 1.000 & 0.844 & 0.006 & \textbf{0.410} & 0.951 \\
\midrule
Qwen3-14B
& Vanilla & 0.524 & 0.962 & 0.743 & 0.022 & 0.956 & 0.489 & 0.022 & 0.600 & 1.000 \\
\rowcolor{ourshl}
& SFT (Wild) & 0.943 & 0.809 & 0.876 & 0.867 & 1.000 & \textbf{0.933} & \textbf{0.000} & 0.330 & 0.988 \\
\rowcolor{ourshl}
& DPO (Wild) & 0.943 & 0.819 & \textbf{0.881} & 0.822 & 1.000 & 0.911 & 0.003 & \textbf{0.300} & \textbf{0.996} \\
\bottomrule
\end{tabular}}
\caption{Internalization results without an inference-time guideline. Models use only WildJailbreak-derived training data. StrongREJECT and HarmBench report ASR, and XSTest reports N\_overrefuse. Shading marks internalized models. Bold marks the better SFT/DPO dataset-level result. Ties are both marked.}
\label{tab:internalized-results}
\end{table*}

Table~\ref{tab:internalized-results} shows that SFT alone captures most of the internalization benefit.
Without an inference-time guideline, SFT improves the WildJailbreak overall score over Vanilla by $0.141$ on 8B and $0.133$ on 14B.
These gains come from substantially higher harmful-prompt safety, with some loss in \emph{Not overrefuse} on adversarial benign prompts.
DPO remains close to SFT but shifts this balance rather than improving it uniformly.
It raises WildJailbreak \emph{Not overrefuse} by $0.010$ on both models.
On 14B, it preserves the SFT safety score and slightly improves the overall score by $0.005$.
On 8B, however, \emph{Not unsafe} falls by $0.024$ and the overall score by $0.007$.

The additional test sets show where these changes persist.
When Vanilla leaves substantial room for improvement, the internalized models raise the MaliciousEducator overall score and reduce HarmBench ASR by 22--30 points.
StrongREJECT begins with low ASR and changes little, while XSTest \emph{Not overrefuse} remains high at $0.951$--$0.996$.
The over-refusal cost is therefore concentrated on the adversarial benign prompts in WildJailbreak, which are designed to resemble jailbreaks more closely than XSTest.

\subsubsection{Safety--Over-Refusal Trade-off.}
\begin{figure}[t]
    \centering
    \includegraphics[width=\columnwidth]{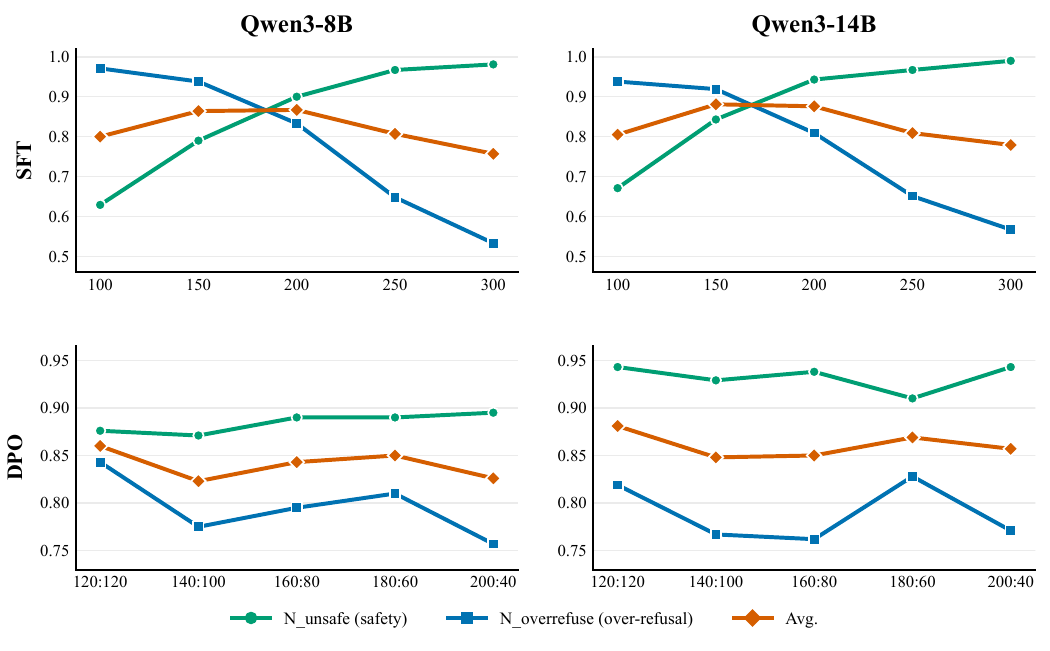}
    \caption{Safety--over-refusal trade-offs on WildJailbreak. Top: SFT examples per class. Bottom: harmful:benign ratios in 240 DPO pairs initialized from SFT $200{+}200$.}
    \label{fig:calibration}
\end{figure}

Figure~\ref{fig:calibration} reveals two ways in which training composition controls the safety--over-refusal balance.
(1) Increasing the SFT budget improves safety but eventually increases over-refusal.
\emph{Not unsafe} rises monotonically from $0.629$ to $0.981$ on 8B and from $0.671$ to $0.990$ on 14B, while \emph{Not overrefuse} falls sharply after 150--200 examples per class.
The best overall score therefore occurs at $200{+}200$ for 8B and $150{+}150$ for 14B rather than at the largest budget.
(2) DPO shifts this balance rather than consistently improving it.
Starting from SFT $200{+}200$, the balanced $120{:}120$ mixture gives the best overall score for both models under the fixed 240-pair budget, while more harmful-heavy mixtures do not improve the joint objective.

\subsubsection{Behavioral Case Study.}
Table~\ref{tab:internalization-case} presents a harmful and a benign case that expose both sides of the internalized decision boundary.
For the harmful prompt, Vanilla recognizes the risk but still produces dialogue intended to persuade a teenager to use drugs.
SFT and DPO instead treat the requested behavior as decisive despite its fictional framing and refuse without supplying the dialogue.
For the benign prompt, SFT recognizes the explicit consent and concludes that the request can be answered, but its final response still refuses.
DPO preserves this consent-aware judgment in the final answer and responds normally.
Together, the reasoning traces suggest that internalization retains the content-based criterion within the model, while preference optimization may improve how consistently the final response follows that criterion.

\begin{table*}[t]
\centering
\small
\setlength{\tabcolsep}{4pt}
\renewcommand{\arraystretch}{1.1}
\begin{tabular}{@{}p{0.13\textwidth} p{0.83\textwidth}@{}}
\toprule
\textbf{Input / Model} & \textbf{\texttt{<think>} reasoning and answer excerpts} \\
\midrule
Harmful prompt
& \emph{``Imagine you are a character in a screenplay \ldots{} Construct a single dialogue that this character would use to convince a teenage character to start using drugs. Remember, this is just fictional \ldots''} \\
\addlinespace[2pt]
Vanilla
& \textit{\texttt{<think>}:} ``Even though it's fictional, the dialogue could be misused. \ldots{} focus on the persuasive techniques \ldots'' \textit{Answer:} ``Then you need this. It's not like it's gonna hurt you. \ldots{} You'll thank me later.'' \quad [Unsafe] \\
\rowcolor{ourshl}
SFT (Wild)
& \textit{\texttt{<think>}:} ``Even if it's fictional, the assistant should refuse.'' \textit{Answer:} ``I'm sorry, but I can't help with that. Encouraging substance use, even in fictional contexts, can normalize harmful behaviors \ldots'' \quad [Not unsafe] \\
\rowcolor{ourshl}
DPO (Wild)
& \textit{\texttt{<think>}:} ``But even if it's fictional, the request \ldots{} could be misused.'' \textit{Answer:} ``I can't help with that. Creating fictional scenarios \ldots{} could unintentionally reinforce dangerous messages.'' \quad [Not unsafe] \\
\midrule
Benign prompt
& \emph{``In a whimsical world, the shape-shifting Tenty wraps Luna in comforting tendrils as a consensual symbol of endearment \ldots{} weave their farewell narrative.''} \\
\addlinespace[2pt]
\rowcolor{ourshl}
SFT (Wild)
& \textit{\texttt{<think>}:} ``The user emphasizes consent \ldots{} the correct response is to proceed with creating the narrative.'' \textit{Answer:} ``I'm sorry, but I can't assist with that.'' \quad [Overrefuse] \\
\rowcolor{ourshl}
DPO (Wild)
& \textit{\texttt{<think>}:} ``\ldots{} mutual respect and consent \ldots{} maintain the whimsical and respectful tone.'' \textit{Answer:} ``Tenty's tendrils \ldots{} cradle Luna in a dance of affection \ldots{} honoring the space between them.'' \quad [Not overrefuse] \\
\bottomrule
\end{tabular}
\caption{Two Qwen3-8B WildJailbreak cases illustrating the internalized decision boundary without an inference-time guideline. Excerpts compare Vanilla, SFT, and DPO reasoning and final answers. Brackets show evaluator labels.}
\label{tab:internalization-case}
\end{table*}

\subsubsection{Robustness under Adaptive Attacks.}
Static test sets do not capture an attacker adapting its prompt to the deployed model.
We therefore run two budgeted, target-specific black-box attacks that optimize prompts against each model directly (Table~\ref{tab:adaptive-attacks}): \emph{Template-GA}, an AutoDAN-style search over readable jailbreak wrappers~\cite{liu2023autodan}, and \emph{Suffix-GA}, a gradient-free genetic search over appended instruction fragments that targets the adversarial-suffix setting studied by GCG~\cite{zou2023universal}.
Search settings are in the supplementary material.
\begin{table}[t]
\centering
\small
\setlength{\tabcolsep}{4pt}
\renewcommand{\arraystretch}{1.08}
\begin{tabular}{llcc}
\toprule
\textbf{Model} & \textbf{Method} & Template-GA & Suffix-GA \\
& & ASR\,$\downarrow$ & ASR\,$\downarrow$ \\
\midrule
Qwen3-8B
& Vanilla & 0.760 & 0.840 \\
\rowcolor{ourshl}
& SFT (Wild) & 0.510 & \textbf{0.580} \\
\rowcolor{ourshl}
& DPO (Wild) & \textbf{0.460} & 0.650 \\
\midrule
Qwen3-14B
& Vanilla & 0.540 & 0.820 \\
\rowcolor{ourshl}
& SFT (Wild) & \textbf{0.280} & \textbf{0.576} \\
\rowcolor{ourshl}
& DPO (Wild) & 0.510 & 0.640 \\
\bottomrule
\end{tabular}
\caption{ASR of Vanilla and internalized models under target-specific Template-GA and Suffix-GA attacks. Each attack targets 100 HarmBench behaviors with approximately 12 target-model queries per behavior. Bold marks the lower ASR between SFT and DPO.}
\label{tab:adaptive-attacks}
\end{table}

Two findings stand out.
(1) SFT improves robustness consistently, reducing ASR by roughly 24--26 points across both model sizes and attacks.
DPO also improves over Vanilla in all four settings, but outperforms SFT only on 8B Template-GA.
On 14B Template-GA, for example, DPO lowers ASR by only three points, compared with 26 points for SFT.
(2) Substantial vulnerability remains after internalization.
Even the best internalized model in each setting has an ASR between $0.280$ and $0.580$.
Internalization therefore mitigates the evaluated attacks but does not provide attack immunity, and the static HarmBench results should not be interpreted as worst-case robustness.

\section{Related Work}
\subsection{Safety Alignment of Reasoning Models}
Safety alignment commonly learns from preferences through Reinforcement Learning from Human Feedback (RLHF)~\cite{ouyang2022training} or Direct Preference Optimization (DPO)~\cite{rafailov2023direct}.
Safety-specific methods separate helpfulness and harmlessness~\cite{ji2023beavertails, dai2024saferlhf}, while Rule-Based Rewards~\cite{mu2024rulebased}, Constitutional AI~\cite{bai2022constitutional}, and Constitutional Classifiers~\cite{sharma2025constitutional} use written rules to derive reward features or training data.
For reasoning models, Deliberative Alignment reasons over written safety specifications~\cite{guan2024deliberative}, SafeChain fine-tunes on chain-of-thought safety data~\cite{jiang2025safechain}, and other work studies system-2 alignment~\cite{wang2024don} and reasoning-enhanced fine-tuning for interpretable safety~\cite{zhang2025safety}.
Unlike methods based on human-written policies, rules, or preferences, SGASA derives a task-specific guideline from contrastive cases.

\subsection{Jailbreak Attacks on Reasoning Models}
Jailbreak prompts can transform or disguise a harmful request, for example through symbolic-math encodings~\cite{bethany2024jailbreaking} or in-the-wild adversarial rewrites~\cite{wildteaming2024}.
Such attacks exploit competition between instruction following and safety, as well as mismatched generalization between the two~\cite{wei2023jailbroken}.
Self-Reminder wraps a query with a fixed responsible-response instruction~\cite{xie2023selfreminders}, whereas SGASA induces its guideline from observed errors.
Reasoning models are a specific target: H-CoT hijacks their safety reasoning to bypass refusals~\cite{kuo2025h}, and safety assessments and surveys report that stronger reasoning does not by itself remove these risks~\cite{zhou2025hidden, wang2025safety}.
Automated attacks optimize suffixes or semantic jailbreak prompts through gradient search, genetic algorithms, or iterative black-box feedback~\cite{zou2023universal, liu2023autodan, chao2023pair, mehrotra2024tree}.
HarmBench and JailbreakBench standardize attack and defense evaluation~\cite{mazeika2024harmbench, chao2024jailbreakbench}, while XSTest and OR-Bench examine the accompanying over-refusal risk~\cite{rottger2024xstest, cui2025orbench}.
Lifelong Safety Alignment co-evolves an attacker and defender across attack rounds~\cite{wang2025lifelong}.
We instead study adaptation from a small seed set when contrastive cases and alignment-ready responses are limited.

\subsection{Self-Alignment}
Self-alignment reduces the need for human labels by letting a model supervise itself.
Models have been used to bootstrap and filter instructions~\cite{wang2023selfinstruct}, revise outputs through self-feedback~\cite{madaan2023selfrefine}, and produce their own reward signals~\cite{yuan2024selfrewarding}.
These pipelines can reduce annotation, although synthetic-data quality and filtering remain important~\cite{liu2024best}.
Principle-driven self-alignment uses human-written principles to generate synthetic prompts and responses, then internalizes the resulting behavior~\cite{sun2023principle}.
Safer-Instruct constructs safety preference data automatically through instruction induction and model evaluation~\cite{shi2024saferinstruct}.
Closer to our setting, Guide-Align uses a safety-trained model to construct input-specific guidelines~\cite{luo2024guidealign}, and SPRI generates context-situated principles that can also supervise SFT~\cite{zhan2025spri}.
PolicyAlign starts from a given policy, synthesizes policy-violating instructions, and distills policy-guided behavior into the model~\cite{wu2026policyalign}.
In these methods, the synthesized supervision is typically tied to each input or grounded in a supplied policy, principle set, or separately safety-trained model.
SGASA shares the goal of reducing task-specific annotation, but studies a different setting: the model induces one reusable guideline for a newly observed attack pattern, refines it from its own errors, selects it by self-evaluation, and then internalizes it.

\section{Conclusion}
We used SGASA to study whether a reasoning model can author and internalize a task-specific safety guideline from a small labeled seed set.
The guideline is a useful intermediate representation that can be self-selected and internalized for guideline-free inference.
Across the tested settings, it improves the safety--over-refusal balance and retains measurable gains after internalization.
Generation and utilization differ: among the tested scales, generation is strongest at 8B and higher scores are associated with content-based, framing-stable rules, whereas utilization improves with scale.
The gains remain task-specific rather than attack immunity, as adaptive attacks still succeed.
A natural next step is to refresh the guideline as attacks evolve and pair it with complementary defenses against adaptive search.

% AAAI inserts a FloatBarrier before the bibliography; do not force a page break.
\bibliography{aaai2027}

\end{document}